\documentclass{dialogue}

\begin{document}

\begin{otherlanguage}{english}
\begin{center}

{\Large\bfseries{Russian Natural Language Generation: Creation of a Language modeling Dataset and Evaluation with Modern Neural Architectures}}

\medskip

Shaheen Z.$^{1,2}$(\texttt{shaheen@itmo.ru}), Wohlgenannt G.$^{1}$ (\texttt{gwohlg@itmo.ru}), \\ Zaity B.$^{2}$ (\texttt{bassel.zaity@gmail.com}), Mouromtsev D. I.$^{1}$ (\texttt{mouromtsev@itmo.ru}),\\Pak V. G.$^{2}$ (\texttt{vadim.pak@cit.icc.spbstu.ru})

\medskip

$^{1}$Faculty of Software Engineering and Computer Systems \\ ITMO University, St. Petersburg, Russia \\
          $^{2}$Institute of Computer Science and Technology \\ Peter the Great St.Petersburg Polytechnic University (SPbPU), \\St. Petersburg, Russia

\end{center}

Generating coherent, grammatically correct, and meaningful text is very challenging, however, it is crucial to many modern
NLP systems. So far, research has mostly focused on English language, for other languages both standardized datasets, as well as experiments with state-of-the-art models, are rare. In this work, we i) provide a novel reference dataset for Russian language modeling, 
ii) experiment with popular modern methods for text generation, namely variational autoencoders, and generative adversarial networks, which we trained on the new dataset. We evaluate the generated text regarding metrics such as perplexity, grammatical correctness and lexical diversity.


\medskip

\end{otherlanguage}

\bigskip

\selectlanguage{english}

\section{Introduction}

Text generation is a key component in many NLP systems that produce text such as translation systems, dialogue systems, or text summarization.
The quality of the generated text is critical in these systems, it should be coherent and well-formed,  
without grammatical mistakes, and semantically meaningful~\cite{guo2018long}. 
Generating human-like text is challenging, it includes modeling high-level syntactic properties and features like sentiment and topic~\cite{Bowman:15}.

Natural Language Generation (NLG) produces human-understandable NL text in a systematic way --
based on non-textual data (eg.~a knowledge base) or from meaning representations (eg.~a given
state of a dialogue system)~\cite{perera2017recent}. Modern NLG systems often make use of (neural) language models~\cite{ruder2019transfer}. 
A language model (LM) is a probability distribution over a sequence of words, and can be used to predict the next word given an input sequence.

In recent years, various types of neural network architectures have been successfully applied in NLG, such as 
variational autoencoders (VAE)~\cite{Bowman:15,Yang:17}, generative adversarial networks (GAN)~\cite{Yu:17,Fedus:18,guo2018long}, and recurrent neural networks (RNN)~\cite{Merity:17}.
Here, we experiment with those architectures on Russian language.

The \emph{goals} of this paper are (i) to create a reference dataset for language modeling for the Russian language, 
comparable to the popular Penn Tree Bank (PTB) dataset for English language, 
and (ii) to adapt and to train several state-of-the-art language models and to evaluate them on the task of Russian language text generation.
We create a dataset of 236K sentences by sampling from the Lenta News dataset, preprocess the text, and filter sentences 
that do not match certain quality criteria. 
Then we train six models (four VAE models with different scheduling methods, seqGAN, and LSTM RNNLM) on the new corpus, 
and evaluate them regarding the perplexity metric, and manually validate 100 sentences for each model regarding grammatical correctness.
We achieve best results with the VAE models, the \emph{zero} variant performs well regarding perplexity, 
but overall the \emph{cyclical} VAE model shows the highest performance, as it generates the largest
fraction of grammatically correct sentences, which have similar characteristics (sentence length, etc.) as the training data. 


\section{Related Work}
\label{sec:rw}

Our Russian language dataset is inspired by the plain-text/language-modeling part of the PTB dataset\footnote{\texttt{https://catalog.ldc.upenn.edu/LDC99T42}}.
PTB contains about 1M words from 1989 Wall Street Journal material, with various annotations such as POS-tags. 
This dataset is very popular among NLP researchers for language modeling and other NLP tasks.
Many recent language models are trained and evaluated also on larger corpora, such as
WikiText-103~\cite{Merity:16}, or WebText~\cite{radford2019language} (created for the GPT transformer models).
For languages other than English high-quality reference datasets are rare.

In language modeling, Recurrent Neural Network Language Models (RNNLM)~\cite{Mikolov:11_b},
and extensions such as long short-term memory (LSTM)~\cite{Sak:14},
are frequently used architectures.
RNNLMs generate text word-by-word depending on a hidden state that summarizes the previous history. 
These models are able to capture long-range dependencies, however, they do not expose interpretable states 
that represent global features like topic or sentiment.
Regarding recent RNNLMs, for example Merity et al.~\cite{Merity:17} investigate different strategies for regularizing and optimizing LSTM-based models. 


Variational Autoencoders (VAEs) \cite{Kingma:13} have been applied to many domains, including language modeling \cite{Bowman:15,Yang:17}. 
They showed impressive results in producing interpretable representations of global features like the topic or of high-level syntactic properties.
For example, Yang et al.~\cite{Yang:17} use the method for unsupervised clustering of the text.
VAEs are trained using regularization to avoid overfitting and produce a regular latent space that has properties enabling the generative process.
Recent research on VAEs focuses on improving the quality of the hidden representation, 
on exploring the properties of the latent space,
and experiments with different architectures to improve VAEs.

Generative adversarial networks GANs~\cite{Goodfellow:14} train a \emph{generator} that tries to produce realistic samples from a data distribution.
The generator is guided by a \emph{discriminator} on how to modify its parameters. 
GANs have been applied successfully to computer vision tasks, however, adapting GANs to generate texts is challenging due to the discrete nature of natural language. Many attempts to adopt GANs to text rely on using reinforcement learning \cite{Yu:17,Fedus:18}, or on Gumbel-Softmax approximation~\cite{Kusner:16} (a continuous approximation of the softmax function).
Zhang et al.~\cite{Zhang:17} use a feature matching scheme for training GANs to generate realistic-looking text.


Little work exists on NLG for the Russian language. Nesterenko~\cite{nesterenko2016building} uses a simple template-based
system to generate stock market news in Russian. Kipyatkova and Karpov~\cite{kipyatkova2017study} study the use of 
RNNLM models in Russian speech recognition. 
Kuratov and Arkhipov~\cite{kuratov2019adaptation} train a BERT (transformer) language model on Russian text (RuBERT)
and evaluate it on tasks such as paraphrase and sentiment detection.
Finally, Shimorina et al.~\cite{shimorina2019webnlg-ru} present an English-Russian parallel corpus for generating 
natural language text from the triples of a knowledge base (data-to-text NLG).
Their corpus was created with neural machine translation.
However, to the best of our knowledge, for general Russian NLG no research work has been published about general-domain NLG datasets and about the evaluation of NLG models based on modern neural architectures.

\section{Variational Autoencoder}
\label{sec:vae}
In this section, we introduce the VAE variants (zero, constant, linear, cyclical) which are applied in the experiments. 
An autoencoder (AE) consists of an encoder that encodes an input sequence into a hidden state and a decoder 
that uses this hidden state to reconstruct the original sequence. 
In a standard AE for language modeling, an RNN is used for both the encoder and the decoder.
The decoder is then used for text generation, where each output token is conditioned 
on the previous output tokens. 
A Variational Autoencoder (VAE) encodes the input sequence $x$ into a region in the latent space rather than a single point, 
this region is defined using a multi-variate Gaussian prior $p(z)$, where the last hidden state of the encoder ($z$) is projected on two separate vectors. These vectors represent the mean and the diagonal co-variance matrix of the prior.
To restore the original sequence, the initial state of the decoder is sampled from the prior, 
and then used to decode the output sequence. This way, the model is forced to be able to decode plausible sentences from every point in the latent space, that has a reasonable probability under the prior~\cite{Bowman:15}. 
A standard recurrent neural network language model is based on a series of next-step predictions, 
thus a standard AE does not provide an interpretable representation of global features such as the topic or of high-level syntactic properties.

The VAE modifies the AE architecture by replacing the deterministic encoder with a learned posterior recognition model $q(z|x)$. 
If the VAE were trained with standard AE reconstruction objective, it would learn to encode $x$ deterministically by making $q(z|x)$ vanishingly small. However, we want the posterior to be close to the prior (most often standard Gaussian), therefore we have two objectives and the goal is to optimize the following lower-bound:
\begin{equation}
    L(\Theta;x) = -KL(q_{\Theta}(z|x)||p(z))+E_{z \sim q_{\theta}(z|x)}[log \, p_{\theta}(x|z)] \leq log \, p(x)
    \label{eq:1}
\end{equation}
The first term is the KL-divergence of the posterior from the prior, and the second is the reconstruction loss, 
where $\theta$ stands for the parameters of the neural network.
Straightforward training of the network using this objective will bring $q(z|x)$ to be exactly the same as the prior $p(z)$, and KL-divergence term in the cost function to zero. 
As a result, the model will behave like a standard RNNLM. Bowman et al.~\cite{Bowman:15} use KL-cost annealing to solve this problem, 
by multiplying a variable weight $\beta$ with the KL term at training time. In the beginning, $\beta$ will be set to zero, 
and then it gets gradually increased, forcing the model to smooth out its encodings and pack them into the prior. 
Later research by Fu et al.~\cite{Liu:19} investigates KL annealing further, they experiment with three scheduling approaches:
\begin{itemize}
\item
Constant Schedule: the standard approach is to keep $\beta = 1$ fixed during training, which causes the vanishing of the KL-term, 
the model will behave as a standard RNNLM.
\item
Monotonic (linear) Annealing Schedule: this is the previously described approach for VAEs by Bowman et al.~\cite{Bowman:15}. 
It starts with $\beta = 0$ and gradually increases during training to $\beta = 1$.  
\item
Cyclical Annealing Schedule: split the training process into $M$ cycles, each cycle consists of two stages:
\begin{enumerate}
\item 
Annealing, where $\beta$ is annealed from 0 to 1 in the first $R [T/M]$ training steps over the cycle; 
$R$: proportion used to increase $\beta$, $T$: the total number of global steps, $M$: the number of cycles.
\item
Fixing: fix $\beta = 1$ for the rest of the cycle. 
\end{enumerate}
\end{itemize}

According to Fu et al. \cite{Liu:19} cyclical KL-annealing results in better latent codes by leveraging the informative representations of previous cycles as warm re-starts.

\section{SeqGAN}
\label{sec:gan}

A Generative Adversarial Network (GAN) contains two main models: the discriminator D is trained to distinguish between real and fake data 
and the generator G, which is trained to fool D and tries to generate samples from the real data distribution. 
D and G are trained simultaneously, the improvement in one model will cause further improvements in the other model. 
When G is able to produce samples from the same data distribution as the real data, D will be no longer able to distinguish 
between real and sampled data.

GANs were applied successfully to tasks in computer vision. However, adapting GANs to text generation is not straightforward, 
because the original GAN works with continuous data, meanwhile, text is discrete and the generator uses a non-differential function 
to produce each token, therefore it is difficult to pass gradient updates from D to G. 
Another difficulty is that D can evaluate only the completed text; for a partially generated sequence
the evaluation score depends on the score of the full sentence.

To address these problems, Yu et al.~\cite{Yu:17} use Reinforcement learning to train GANs for text generation. 
The generator is treated as an agent of reinforcement learning, and the state corresponds to the generated tokens up to this moment 
and the action is the next token to be generated. A discriminator will give the feedback reward that guides the learning process.

We can consider G as a stochastic parametric policy where Monte Carlo (MC) search is used to approximate the state-action value using the policy gradient. 
D is trained by providing positive examples from the real data and negative examples from the generated samples by G. 
G is updated using the policy gradient and MC search by the reward signal received from D. 
The reward represents how likely the discriminator is fooled by the generator. 
Yu et al.~\cite{Yu:17} use a rollout policy with MC search, they set the rollout policy the same as G during the experiments. 
For G they choose a recurrent neural network with LSTM cells, and D uses a convolutional neural network with highway architecture. 

The training strategy is as follows: first G is pre-trained using maximum likelihood estimation, then G and D are trained alternatively. 
To train D, at each training step we sample negative examples using G and use examples from the true data as the positive examples.

\section{Dataset}
\label{sec:dataset}

\begin{figure}
  \centering
  \begin{tabular}[b]{c}
    \includegraphics[height=3in, width=.6\linewidth]{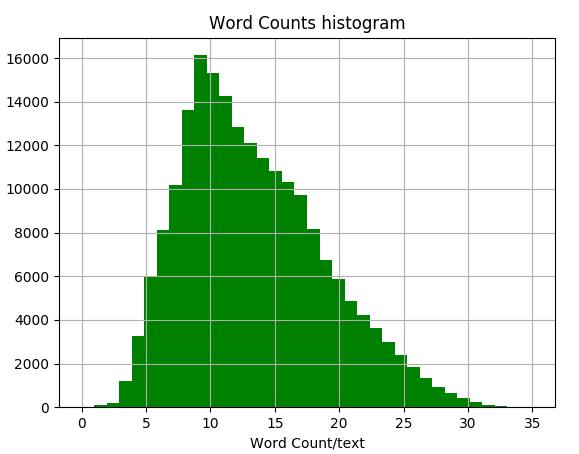} r\\
    \small (a)
  \end{tabular} \qquad
  \begin{tabular}[b]{c}
    \includegraphics[width=.25\linewidth]{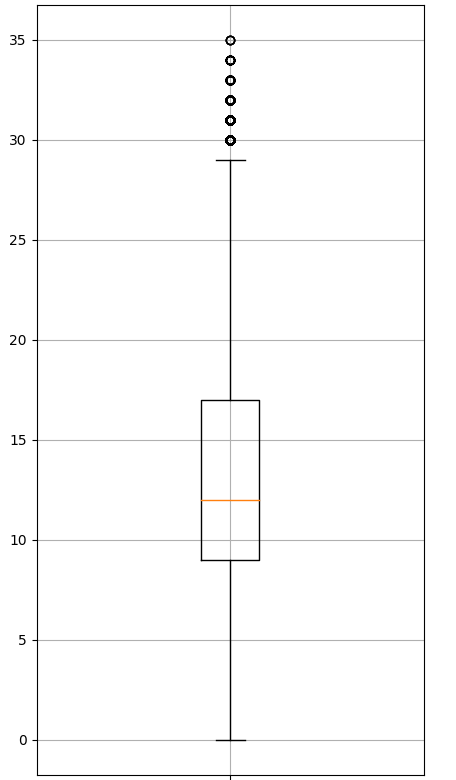} \\
    \small (b)
  \end{tabular}
  \caption{a) A histogram visualizing the number of words per sentence which shows the probability distribution of the data. 
b) A box plot for the number of words per sentence.}
\label{fig1}
\end{figure}

In this section we discuss the Russian language dataset for language modeling, its creation, and characteristics.

Penn Tree Bank (PTB)~\cite{Mikolov:11_a} is a very popular dataset for experimenting with language models, 
and many researchers use it in experiments with the same settings (dataset splits, etc.) -- 
which allows to compare different language modeling approaches. 
PTB is heavily preprocessed, and the dataset vocabulary is limited to 10,000 tokens with no numbers, punctuation or capital letters.

Our goal was to create a similar reference dataset for the Russian language.
Our dataset is based on the Lenta news dataset\footnote{\texttt{https://github.com/yutkin/Lenta.Ru-News-Dataset}}, 
a corpus of over 800K Russian news articles collected from Lenta.Ru between 1999--2019. 
To create our dataset we randomly sample sentences from the Lenta dataset after we apply preprocessing in a similar way as in PTB.

The \emph{preprocessing pipeline} includes: a) lower-casing the text, b) replacing all URLs with a special token <url>, 
c) separating punctuation symbols from neighboring tokens with spaces, 
and finally d) replacing digits with a special character \emph{D}, and for any 5 or more consecutive digits we use the special character \emph{N}. Therefore, numbers in the dataset will take the following forms: \emph{D, DD, DDD, DDDD, or N}. 
Step d) helps to keep meaningful numeric tokens, such as in these examples:
\begin{itemize}
    \item 
    часть D статьи DDD ук
    \item
    встреча так и завершилась со счетом D : D
    \item
    в DDDD году, госкорпорация обеспечивает DD процента электроэнергии .
\end{itemize}

Subsequently, we apply the NLTK\footnote{\texttt{https://nltk.org}}
PunktSentenceTokenizer\footnote{\texttt{https://github.com/Mottl/ru\_punkt}}
to tokenize the text into sentences. We create a dictionary using the 15,000 most frequent tokens, and
replace all other tokens in the text with <UNK>.
In contrast to PTB, which uses a dictionary size of 10,000, we decided to keep 15,000 tokens because of the rich morphology 
of the Russian language where nouns, adjectives, and verbs change forms according to their role in the sentence.

Finally, we create the Russian language corpus by sampling 200,000 examples for the training set, 
16,000 for the development set and 16,000 for the test set. In the sampling process, we only accepted sentences 
which fulfill the following conditions:
\begin{itemize}
    \item  The sentence contains less than 40 tokens. 
    \item  The sentence does not include any English words. 
    \item  The number of single (') and double (") quotation marks is even (balanced).
    \item  Every opening bracket is followed by a closing bracket (balancing condition).
    \item  Less than 10\% of the tokens in the sampled sentence are the <UNK> token.
\end{itemize}

The dataset and code are available on GitHub\footnote{\texttt{https://github.com/zeinsh/lenta\_short\_sentences}}.
Finally, we provide an overview of dataset statistics, including the mean number and standard deviation of the 
number of tokens per sentence in Table~\ref{tab1}. 
The histogram and the box plot in Fig.~\ref{fig1} show the distribution of sentence length (number of tokens) in the dataset.
Those statistics will be helpful to compare the sentences generated with various methods (see next section) with
the original dataset.

\begin{table}[htbp]
\centering
\caption{An overview of the Russian language modeling corpus (statistics of training, development and test set).}
\label{tab1}
\begin{tabular}{|c|c|c|c|}\hline
  & Training set & Development set & Test set \\\hline
\#examples & 200000 & 16000 & 16000 \\\hline
Mean \#tokens per example & 13.26 & 13.17 & 13.14 \\\hline
Stddev for \#tokens per example & 5.63 & 5.66 & 5.62 \\\hline
\#unique tokens & 14511 & 13398 & 13401 \\\hline
\end{tabular}
\end{table}



\section{Experiments}
\label{sec:eval}

\subsection{Evaluation Setup}

We experiment with the following popular text generation methods:
VAE, GAN, and RNN.

For the \textbf{VAE} method, we build on the implementation by Baumgärtner\footnote{\texttt{https://github.com/timbmg/Sentence-VAE}} and add the implementation of additional scheduling methods, namely cyclical, constant and zero scheduling. 
Constant and cyclical scheduling are explained in Section~\ref{sec:vae}, 
and in the zero schedule we set $\beta = 0$, 
which excludes the KL-divergence term from the lower-bound computation in equation~\ref{eq:1}.
For the VAE encoder, we use 300-dim input embeddings, a single layer of 256 LSTM cells, and
a latent vector (bottleneck) size of 16, and in the decoder again 256 LSTM cells and a 300-dim output. We train the VAEs for 10 iterations with an embedding dropout probability of 0.5.

The experiments with \textbf{seqGAN} are based on the PyTorch implementation on GitHub\footnote{\texttt{https://github.com/suragnair/seqGAN}}, 
We modified the original implementation to adapt it to our text dataset instead of discrete numbers created by the oracle generator in the original implementation.
For seqGAN, we also use 300-dim input embeddings, and a single layer of 256 LSTM cells.
We pretrain G for 10 iterations with an embedding dropout probability of 0.5, 
and then pretrain D (10 iterations). 
Then follow 10 epochs of adversarial training, each of which trains G for one iteration 
and D for five iterations.
Finally, to measure the effect of the adversarial training on the RNNLM, 
we performed an evaluation on the pre-trained LSTM generator separately as \textbf{RNNLM} 
(see Section \ref{sec:eval_results}).


For evaluation, we use a dual strategy. First, in line with most research on language modeling,
we calculate the \emph{perplexity} on the test set for each model.
Perplexity is a measure of how well a probability distribution predicts a sample.
Perplexity does not always correlate with human correlation, in fact there is sometimes 
a negative correlation~\cite{chang2009reading}, for this reason we also include an human expert evaluation (see below).

Furthermore, we generate 10,000 sample texts with each method by greedily sampling word by word.
Examples of the results are given in GitHub repository\footnote{\texttt{https://github.com/zeinsh/lenta\_short\_sentences/blob/master/samples.md}} and in Appendix 1.
For the generated samples we use expert evaluation, 
where a Russian native speaker checks the generated text for grammatical correctness, 
and assigns a score of either $1$ or $0$. The value $1$ signifies that no grammatical mistakes were found in the text. As we manually evaluate 100 unique sentences for each model, the maximum score per model is 100.

\subsection{Evaluation Result}
\label{sec:eval_results}

Here we analyze the models with regards to the following aspects: perplexity, token statistics of generated text,
and manual evaluation of grammatical correctness. Furthermore, for VAE models we discuss the spatial distribution of latent representations.


In Table~\ref{tab2} we report on the perplexity metric for the test set. The \emph{zero}
VAE model clearly shows the best results, followed by the cyclical model. As we will see in Tables~\ref{tab3} and
~\ref{tab4}, despite good results on perplexity, the \emph{zero} model does not excel in grammatical correctness of the generated text.

\begin{table}[htbp]
\centering
\caption{Perplexity calculated on test set for each model.}
\label{tab2}
\begin{tabular}{|c|c|c|c|c|c|c|}\hline
x  & zero & constant & linear & cyclical & RNNLM & seqGAN \\\hline
perplexity & 7.19 & 16.27 & 14.36 & 14.11 & 27.88 & 27.93 \\\hline

\end{tabular}
\end{table}

Table~\ref{tab3} gives an overview of some statistics of the generated text samples.
We can see that LSTM RNNLM, seqGAN, and especially the \emph{zero} variant of VAE produce 
a large number of unique sentences -- with very little overlap with the training 
sentences. On the other hand, the \emph{constant} VAE model fails to generate a large variety
of sentences. Liu et al.~\cite{Liu:19} argue that the constant schedule ignores $z$ and treats it as noise. 
Regarding sentence length, most models are similar, except \emph{constant}, which creates shorter sentences. 
More interestingly, the \emph{zero} and \emph{constant} model show little variance in sentence length, 
while the other models much better capture the variance in sentence length of the training dataset. 
Finally, only RNNLM and seqGAN are able to have a diversity in vocabulary similar to the training data.

\begin{table}[htbp]
\centering
\caption{Comparison between models regarding uniqueness and sentence length of the generated data.}
\label{tab3}
\begin{tabular}{|c|c|c|c|c|c|c|}\hline
  & zero & constant & linear & cyclical & RNNLM & seqGAN \\\hline
\# unique sent.~(out of 10000)   & 10000 & 231 & 8810 & 8868 & 9972 & 9979\\\hline
\# unique sent.~not in train-set & 10000 & 225 & 8694 & 8752& 9972 & 9979\\\hline
Mean \#tokens per sample&10.52&8.60&11.89&11.25&13.53&13.89\\\hline
Std-dev of \# tokens per sample&2.25&2.27&5.08&4.78&5.99&6.08\\\hline
\# unique words&5324&353&5028&4550&11430&11505\\\hline
\end{tabular}
\end{table}

As mentioned, a Russian native speaker manually verified 100 generated sentences for each model
regarding grammar. The results (number of grammatically correct sentences) are 
presented in Table~\ref{tab4}. The data clearly shows that the \emph{cyclical} VAE model 
performs very well (91\% correct sentences), while the \emph{zero} VAE model, although providing low perplexity, produces many grammatically wrong sentences. This corresponds with Chang et al.~\cite{chang2009reading}, ie.~that perplexity does not always correlate with human evaluation. In our experiments, RNNLM and seqGAN fail to 
generate a high ratio of grammatically correct sentences.

\begin{table}[htbp]
\centering
\caption{Number of grammatically correct sentences (out of 100) checked by a native speaker.}
\label{tab4}
\begin{tabular}{|c|c|c|c|c|c|c|}\hline
  & zero & constant & linear & cyclical & RNNLM & seqGAN \\\hline
Score & 77 & 79 & 86 & 91 & 43 & 51 \\\hline
\end{tabular}
\end{table}

\begin{figure}[htbp]
\centering
\includegraphics[scale=.4]{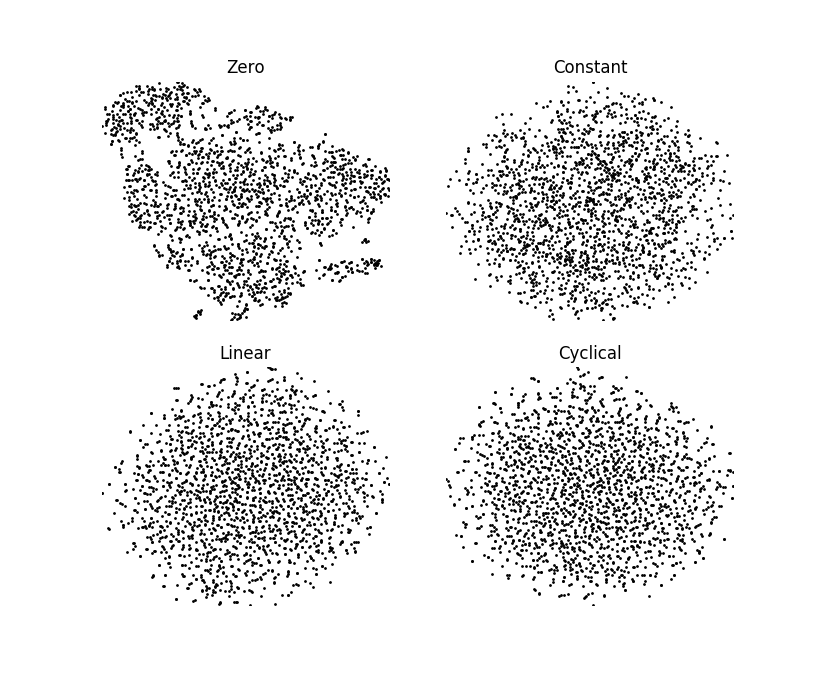}
\caption{Latent representation of texts in the development set. The KL-divergence term in the constant, linear, and cyclical schedule forces the encoder to fill the latent space.}
\label{fig2}
\end{figure}

Fig.~\ref{fig2} shows a projection of the latent representations in the development set into 2D space (using tSNE\footnote{\texttt{https://scikit-learn.org/stable/modules/generated/sklearn.manifold.TSNE.html}}).
The figure compares the resulting distributions of the VAE methods with a zero, constant, linear and cyclical schedule.
The figure shows that the zero schedule, which corresponds to a standard autoencoder, produces a rather irregular  
distribution of latent codes. The KL-divergence term in VAEs causes the algorithm to fill the latent space. \\
As discussed before, although zero gives the best perplexity on the test set, samples from the zero model contain many grammatically incorrect sentences, as it sometimes samples $z$ from regions in the latent space with low density. That explains why samples from linear and cyclical VAES are better in terms of grammar, where latent codes produced by these models fill the latent space. 

Finally, in the appendix and in the GitHub repository\footnote{\texttt{https://github.com/zeinsh/lenta\_short\_sentences/blob/master/interpolation.md}}, 
we give examples on how VAE models can interpolate 
between two sentences. 
Following Bowman et al.~\cite{Bowman:15}, we sample two random points in the latent space 
and then decode those into two sentences. 
Then, starting from the first sentence, we gradually move through the latent space on a line
to the second sentence, and pick points on the way, which are decoded into sentences.
This process makes the ability of interpolation in the latent space explicit.

In summary, we found that the cyclical VAE produces best result with regards to grammatical correctness, 
followed by the VAE with linear schedule. Both also generate sentences with similar characteristics as in the training set (regarding sentence length), however concerning sentence length and diversity of the used vocabulary 
the plain RNNLM and the seqGAN produce better results. But RNNLM and seqGAN generate a high number of grammatically wrong sentences in the experiments.

\section{Conclusion}
\label{sec:concl}

In this work, we present a new dataset for Russian language modeling (based on the Lenta News dataset) 
and perform a comparative study between two modern methods in text generation, namely VAE and seqGAN.
Our results show the effect of the scheduling method on the quality of the generated text in VAEs, 
where linear and cyclical schedules produced the best models grammatically, however, 
the zero method showed the best perplexity, but an irregular distribution of the latent codes. 
LSTM and SeqGAN were able to replicate the mean and variance of the length of sentences in the original dataset as well as the number of unique words.
The contributions of this work are: i) the provision (on GitHub) of a reference dataset for Russian language modeling with 236K sentences in total, ii) the adaption of various VAE variants and seqGAN to Russian text, 
iii) and extensive experiments and evaluations with the chosen deep learning methods which indicate that the cyclical VAE approach performs best overall.
Future work will include a deeper investigation of the latent representations produced by VAEs (and why VAEs produce less diverse sentences), apply state-of-the-art models like LeakGAN and studying the generation of Russian language text 
conditioned on topic, sentiment, etc.

\section*{Acknowledgments}
This work was supported by the Government of the Russian Federation (Grant 074-U01) through the ITMO Fellowship and Professorship Program.

\bibliography{dialogue}

\begin{thebibliography}{10}

\bibitem{Bowman:15}
Samuel~R Bowman, Luke Vilnis, Oriol Vinyals, Andrew~M Dai, Rafal Jozefowicz,
  and Samy Bengio.
\newblock {Generating sentences from a continuous space}.
\newblock {\em arXiv preprint arXiv:1511.06349}, 2015.

\bibitem{chang2009reading}
Jonathan Chang, Sean Gerrish, Chong Wang, Jordan~L Boyd-Graber, and David~M
  Blei.
\newblock Reading tea leaves: How humans interpret topic models.
\newblock In {\em Advances in neural information processing systems}, pages
  288--296, 2009.

\bibitem{Fedus:18}
William Fedus, Ian Goodfellow, and Andrew~M Dai.
\newblock {Maskgan: Better text generation via filling in the {\_}}.
\newblock {\em arXiv preprint arXiv:1801.07736}, 2018.

\bibitem{Goodfellow:14}
Ian Goodfellow, Jean Pouget-Abadie, Mehdi Mirza, Bing Xu, David Warde-Farley,
  Sherjil Ozair, Aaron Courville, and Yoshua Bengio.
\newblock {Generative adversarial nets}.
\newblock In {\em Advances in neural information processing systems}, pages
  2672--2680, 2014.

\bibitem{guo2018long}
Jiaxian Guo, Sidi Lu, Han Cai, Weinan Zhang, Yong Yu, and Jun Wang.
\newblock Long text generation via adversarial training with leaked
  information.
\newblock In {\em Thirty-Second AAAI Conference on Artificial Intelligence},
  2018.

\bibitem{Kingma:13}
Diederik~P Kingma and Max Welling.
\newblock {Auto-encoding variational bayes}.
\newblock {\em arXiv preprint arXiv:1312.6114}, 2013.

\bibitem{kipyatkova2017study}
Irina~S Kipyatkova and Alexey~A Karpov.
\newblock A study of neural network russian language models for automatic
  continuous speech recognition systems.
\newblock {\em Automation and Remote Control}, 78(5):858--867, 2017.

\bibitem{kuratov2019adaptation}
Yuri Kuratov and Mikhail Arkhipov.
\newblock Adaptation of deep bidirectional multilingual transformers for
  russian language.
\newblock {\em arXiv preprint arXiv:1905.07213}, 2019.

\bibitem{Kusner:16}
Matt~J Kusner and Jos{\'{e}}~Miguel Hern{\'{a}}ndez-Lobato.
\newblock {Gans for sequences of discrete elements with the gumbel-softmax
  distribution}.
\newblock {\em arXiv preprint arXiv:1611.04051}, 2016.

\bibitem{Liu:19}
Xiaodong Liu, Jianfeng Gao, Asli Celikyilmaz, and Lawrence Carin.
\newblock {Cyclical annealing schedule: A simple approach to mitigating kl
  vanishing}.
\newblock {\em arXiv preprint arXiv:1903.10145}, 2019.

\bibitem{Merity:17}
Stephen Merity, Nitish~Shirish Keskar, and Richard Socher.
\newblock {Regularizing and optimizing LSTM language models}.
\newblock {\em arXiv preprint arXiv:1708.02182}, 2017.

\bibitem{Merity:16}
Stephen Merity, Caiming Xiong, James Bradbury, and Richard Socher.
\newblock {Pointer sentinel mixture models}.
\newblock {\em arXiv preprint arXiv:1609.07843}, 2016.

\bibitem{Mikolov:11_a}
Tom{\'{a}}{\v{s}} Mikolov, Anoop Deoras, Stefan Kombrink, Luk{\'{a}}{\v{s}}
  Burget, and Jan {\v{C}}ernock{\'{y}}.
\newblock {Empirical evaluation and combination of advanced language modeling
  techniques}.
\newblock In {\em 12th Annual Conf.~of the International Speech Communication
  Association}, 2011.

\bibitem{Mikolov:11_b}
Tom{\'{a}}{\v{s}} Mikolov, Stefan Kombrink, Luk{\'{a}}{\v{s}} Burget, Jan
  {\v{C}}ernock{\'{y}}, and Sanjeev Khudanpur.
\newblock {Extensions of recurrent neural network language model}.
\newblock In {\em 2011 IEEE intern. conf.~on acoustics, speech and signal
  proc.~(ICASSP)}, pages 5528--5531. IEEE, 2011.

\bibitem{nesterenko2016building}
Liubov Nesterenko.
\newblock Building a system for stock news generation in russian.
\newblock In {\em Proc. of 2nd Int. Workshop on NLG and the Semantic Web
  (WebNLG 2016)}, pages 37--40, 2016.

\bibitem{perera2017recent}
Rivindu Perera and Parma Nand.
\newblock Recent advances in natural language generation: A survey and
  classification of the empirical literature.
\newblock {\em Computing and Informatics}, 36(1):1--32, 2017.

\bibitem{radford2019language}
Alec Radford, Jeffrey Wu, Rewon Child, David Luan, Dario Amodei, and Ilya
  Sutskever.
\newblock Language models are u/nsupervised multitask learners.
\newblock {\em OpenAI Blog}, 1(8):9, 2019.

\bibitem{ruder2019transfer}
Sebastian Ruder, Matthew~E Peters, Swabha Swayamdipta, and Thomas Wolf.
\newblock Transfer learning in natural language processing.
\newblock In {\em Proc. of 2019 Conf. of the North American Chapter of the
  Association for Computational Linguistics: Tutorials}, pages 15--18, 2019.

\bibitem{Sak:14}
Hasim Sak, Andrew~W Senior, and F.~Beaufays.
\newblock {Long short-term memory recurrent neural network architectures for
  large scale acoustic modeling}.
\newblock {\em research.google}, 2014.

\bibitem{shimorina2019webnlg-ru}
Anastasia Shimorina, Elena Khasanova, and Claire Gardent.
\newblock Creating a corpus for {R}ussian data-to-text generation using neural
  machine translation and post-editing.
\newblock In {\em Proceedings of the 7th Workshop on Balto-Slavic Natural
  Language Processing}, pages 44--49, Florence, Italy, August 2019. Association
  for Computational Linguistics.

\bibitem{Yang:17}
Zichao Yang, Zhiting Hu, Ruslan Salakhutdinov, and Taylor Berg-Kirkpatrick.
\newblock {Improved variational autoencoders for text modeling using dilated
  convolutions}.
\newblock In {\em Proc. 34th Int. Conf. on Machine Learning-Volume 70}, pages
  3881--3890. JMLR. org, 2017.

\bibitem{Yu:17}
Lantao Yu, Weinan Zhang, Jun Wang, and Yong Yu.
\newblock {Seqgan: Sequence generative adversarial nets with policy gradient}.
\newblock In {\em 31st AAAI Conf. on Art. Intelligence}, 2017.

\bibitem{Zhang:17}
Yizhe Zhang, Zhe Gan, Kai Fan, Zhi Chen, Ricardo Henao, Dinghan Shen, and
  Lawrence Carin.
\newblock {Adversarial feature matching for text generation}.
\newblock In {\em Proc. 34th Int. Conference on Machine Learning-Volume 70},
  pages 4006--4015. JMLR. org, 2017.

\end{thebibliography}

\clearpage

\section*{Appendix 1}

In the appendix, we show the interpolation using the four VAE models trained on the reference dataset. 
As described in Section~\ref{sec:eval_results} we sample two random points in the latent space 
and then decode those into two sentences. 
Starting from the first sentence, we gradually move through the latent space on a line
to the second sentence, and pick points on the way, which are decoded into sentences.
This processes makes the ability of interpolation in the latent space explicit. \\
\\
\\
\textbf{VAE/zero schedule} 
\\
\\-  пожары с ними возник конфликт между двумя группами и на юго - востоке <unk> . <eos>
\\-  пожары с ними возник конфликт между двумя группами и на юго - востоке <unk> . <eos>
\\-  пожары с рельсов сошел с одной из самых опасных технологий , вызванных <unk> на работу . <eos>
\\-  пожары с созданием по атомной энергии ( магатэ ) , на <unk> островах . <eos>
\\-  пожары с созданием по спасению ракет у ворот у берегов острова , <unk> за рубеж . <eos>
\\-  соперник по многим показателям проходит у дома от продажи билетов , <unk> за рубеж . <eos>
\\-  соперник по многим показателям ( нсн ) создает у нее , <unk> за рубеж . <eos>
\\-  соперник ( по словам представителя рфс ) при этом <unk> , сообщает тасс . <eos>
\\-  соперник ( по словам ) у него возникли трудности , <unk> за собой . <eos>
\\-  соперник - гран - при россии по легкой атлетике , <unk> за рубеж . <eos>\\
\\
\textbf{VAE/constant schedule} \\
Explanation: This model decoded both points from the latent space into the \emph{same} sentence.
\\
\\-  в dddd году он был объявлен в международный розыск . <eos>
\\-  в dddd году он был объявлен в международный розыск . <eos>
\\-  в dddd году он был объявлен в международный розыск . <eos>
\\-  в dddd году он был объявлен в международный розыск . <eos>
\\-  в dddd году он был объявлен в международный розыск . <eos>
\\-  в dddd году он был объявлен в международный розыск . <eos>
\\-  в dddd году он был объявлен в международный розыск . <eos>
\\-  в dddd году он был объявлен в международный розыск . <eos>
\\-  в dddd году он был объявлен в международный розыск . <eos>\\
\\
\textbf{VAE/linear schedule} 
\\
\\-  точная дата выхода фильма пока неизвестна , пока неясно , выйдет на экраны . <eos>
\\-  точная дата выхода фильма пока неизвестна , пока неясно , не уточняется . <eos>
\\-  такое заявление сделал на заседании совета федерации хоккея россии , передает риа новости . <eos>
\\-  соответствующее заявление сделал на заседании совета федерации хоккея россии по футболу , передает риа новости . <eos>
\\-  соответствующее заявление сделал на заседании совета федерации по правам человека , передает риа новости . <eos>
\\-  соответствующее заявление сделал на заседании совета федерации по правам человека , передает риа новости . <eos>
\\-  соответствующее заявление сделал в четверг , dd мая , на сайте следственного комитета рф . <eos>
\\-  соответствующее заявление сделал в четверг , dd мая , в ходе совещания . <eos>
\\-  соответствующее заявление в четверг , dd мая , приводит риа новости . <eos>
\\-  соответствующее постановление опубликовано на сайте ведомства в четверг , dd мая , в <unk> . <eos>\\
\\
\textbf{VAE/cyclical schedule} 
\\
\\-  в то же время , по словам <unk> , он посетит россию , а также в последние годы . <eos>
\\-  в то же время , по словам <unk> , он посетит россию , а также в последние годы . <eos>
\\-  в то же время , по словам <unk> , он был вынужден уйти в отставку . <eos>
\\-  в <unk> , где он жил в нью - йорке , не уточняется . <eos>
\\-  в <unk> , где он жил в нью - йорке , не уточняется . <eos>
\\-  в результате <unk> погибли dd человек , большинство из которых были убиты . <eos>
\\-  по его словам , <unk> был задержан в ходе проверки , проведенной полицией . <eos>
\\-  по его словам , <unk> был задержан в ходе перестрелки с полицией . <eos>
\\-  по предварительным данным , <unk> был ранен , а также ранен . <eos>
\\-  причины катастрофы устанавливаются , <unk> в результате инцидента никто не пострадал . <eos>\\

\section*{Appendix 2}

Here we present a few examples of grammatically correct and incorrect sentences generated by the VAE/Cyclical model.
\\
\\
\textbf{Grammatically correct}
\\
\\ - если вина будет доказана , <unk> грозит до dd лет лишения свободы . <eos>
\\ - но это не первый случай , когда он будет <unk> в прямом эфире . <eos>
\\ - в последние годы он жил в нью - йорке и вашингтоне . <eos>
\\ - стоимость контракта оценивается в dd миллионов долларов . <eos>
\\
\\
\textbf{Grammatically incorrect}
\\
\\ - если бы не удастся , то <unk> , как , будет <unk> , то , как он , не будет делать какие - то проблемы , касающиеся <unk> изменений в закон " о <unk> " . <eos>
\\ - поединок состоялся в ночь на d февраля , однако известно , что <unk> в нем принял участие около dd тысяч человек . <eos>
\end{document}


Samples Generated using each model in the experiments

VAE/zero
судя по всему миру \— восстановление \<unk\> от него получается . \<eos\>
число рабочих мест по всему миру , проживающих за пределами россии . \<eos\>
пожары приступили к катастрофе у мемориала управления делами президента россии . \<eos\>
мины недалеко от от морского флота , dd - летняя девушка . \<eos\>
главное управление по противодействию коррупции ( « интерфакс » , запрещена на украине . \<eos\>
незадолго до начала войны у нас есть шансы разрешения властей , \<unk\> на родину в европу . \<eos\>
пресс - служба регионального управления скр по республике , \<unk\> за собой право . \<eos\>
уполномоченный по правам человека ( вкс ) россии на юге страны . \<eos\>
взаимодействие - членов альянса по развитию гражданского населения , так и \<unk\> не \<unk\> . \<eos\>
судя по всему миру приходится от тренировок из - \<unk\> , проживающих под контролем . \<eos\>
судя по всему миру , у нас есть шансы с путиным « \<unk\> » . \<eos\>

VAE/constant
в результате взрыва погибли dd человек , еще dd получили ранения . \<eos\>
в dddd году в россии \<unk\> dd процентов россиян , а также \<unk\> сборы картины . \<eos\>
в dddd году в россии были введены ограничения на въезд в страну . \<eos\>
в dddd году он был объявлен в международный розыск . \<eos\>
об этом сообщает « интерфакс » . \<eos\>
в dddd году в россии были введены ограничения на курение в \<unk\> . \<eos\>
в результате аварии погибли dd человек , еще dd получили ранения . \<eos\>
в результате аварии погибли dd человек , еще dd получили ранения . \<eos\>
об этом сообщает " интерфакс " . \<eos\>
в dddd году в россии в dddd году \<unk\> в dddd году , а в dddd году \— ddd , d миллиона долларов . \<eos\>

VAE/Linear
как передает риа новости со ссылкой на источник в правоохранительных органах , \<unk\> \<unk\> в соцсетях , а также в целях обеспечения безопасности дорожного движения . \<eos\>
в dddd году в россии пройдет dd \<unk\> . \<eos\>
представители сша заявили , что не признает вины \<unk\> , а также не \<unk\> . \<eos\>
« \<unk\> » запустила на базе одноименного управления по борьбе с организованной преступностью и \<unk\> . \<eos\>
по итогам прошлого года экономика страны сократится на d , d процента до dddd , d пункта , а импорт - d , d процента . \<eos\>
в частности , в dddd году , когда он был вынужден совершить посадку , а затем \<unk\> \<unk\> . \<eos\>
в результате инцидента никто не пострадал . \<eos\>
в dddd году в \<unk\> были убиты более ddd тысяч человек . \<eos\>
в настоящее время на месте происшествия работают специалисты мчс россии и \<unk\> , \<unk\> на dd процентов . \<eos\>
таким образом , \<unk\> пропустит матч за счет \<unk\> , а также в финале лиги чемпионов . \<eos\>

VAE/cyclic
в dddd году в россии было продано более ddd миллионов долларов . \<eos\>
в \<unk\> автономном округе москвы , по данным следствия , dd марта dddd года , в результате которого погибли dd человек , dd получили ранения . \<eos\>
однако , по мнению аналитиков , это приведет к сокращению дефицита бюджета . \<eos\>
как сообщает риа новости , в среду , dd марта , президент россии владимир путин подписал указ о введении санкций против россии , \<unk\> в отношении россии . \<eos\>
однако , по мнению главы государства , необходимо , чтобы избежать этого проекта , \<unk\> , что , возможно , будет \<unk\> в течение ближайших недель . \<eos\>
на месте происшествия работают сотрудники милиции , двое из них находятся в тяжелом состоянии . \<eos\>
в результате погибли dd человек . \<eos\>
об этом сообщает « интерфакс » . \<eos\>
поединок состоялся в ночь на d февраля , однако известно , что \<unk\> в нем приняли участие около dd тысяч человек . \<eos\>
\<unk\> также отметил , что в настоящее время в данном районе находятся около dd тысяч человек . \<eos\>

RNNLM
один пакет сообщили , что украина запрещена поставка \<unk\> оценивается в D тысяч долларов , сообщает " коммерсант " со счетом D на D .
« все же надо заработали от D , D миллиарда долларов .
в результате теракта начались службу DD , D \<unk\> DDD тысяч человек .
« по мнению сирийского агентства " , в результате пожар погибли ранены двое международному человеческих подписчиков .
\<unk\> пострадавших не считаются в у двух стран , а \<unk\> пока года назад является \<unk\> оружием в течение DD дней , по его словам , юридических лиц на борту .
по предварительным данным чеченских стороны , в том числе мчс , тот , кто имеет право самой техники .
\<unk\> был составлен на ее первый место по состоянию на DD миллиардов евро .
также на рассмотрение палаты по который оказывает \<unk\> офиса сообщил было \<unk\> .
в « аналогичный активе DDD \<unk\> по услугами мирового пространства .
кроме того , вместе с тем известно , что инфляция в минобороны сша в DDDD году было возбуждено по цене пять DDD , D тысяч долларов ) .

seqGAN
эта картина было " официальное " .
о \<unk\> уходе не , что касается .
" заочно мы не не - " законодательного органа президента " михайлов и \<unk\> на были выставлены пост .
" прекращено " сериала " \<unk\> .
тогда они сообщили , что мы принял один из поста полиции о том , что касается .
однако данную рекламу на востоке \<unk\> магазин специалисты отказались выйти на след его ведения видимости .
служба с студийных москвы .
кроме того , что все эти обвинения в причинении смерти по неосторожности данным \<unk\> в настоящий момент .
DD июля число \<unk\> был DD апреля .
президент была будет выступить в результате , однако освободить человека в связи с комиссар евросоюза .


\begin{figure}%
    \centering
    \subfloat[label 1]{{\includegraphics[width=5cm]{img1} }}%
    \qquad
    \subfloat[label 2]{{\includegraphics[width=5cm]{img2} }}%
    \caption{2 Figures side by side}%
    \label{fig:example}%
\end{figure}